%% file: main.tex
\documentclass[11pt]{article}

\usepackage[final]{acl}

\usepackage{times}
\usepackage{latexsym}

\usepackage[T1]{fontenc}

\usepackage[utf8]{inputenc}

\usepackage{microtype}

\usepackage{inconsolata}

\usepackage{graphicx}

\usepackage{amsmath} 
\usepackage{amssymb} 
\usepackage{booktabs}
\usepackage{multirow}
\usepackage{enumitem}
\usepackage{hyperref}
\usepackage{xurl}
\usepackage{amsthm}
\usepackage{threeparttable}

\usepackage{listings}
\usepackage{xcolor}
\usepackage{array}
\usepackage{tcolorbox}
\tcbuselibrary{skins}

\usepackage{fontawesome5}
\usepackage{booktabs}
\usepackage{amssymb}
\usepackage{xcolor}

\definecolor{sepseq_green}{HTML}{0B6E4F} 
\definecolor{vanilla_grey}{HTML}{6E6E6E} 

\lstdefinelanguage{json}{
    string=[s]{"}{"},
    stringstyle=\color{-},
    comment=[l]{//},
    morecomment=[s]{/*}{*/},
    commentstyle=\color{gray}\itshape,
    numbers=left,
    numberstyle=\tiny,
    stepnumber=1,
    numbersep=8pt,
    showstringspaces=false,
    breaklines=true,
    frame=lines,
    backgroundcolor=\color{gray!10},
    literate=
     *{0}{{{\color{+}0}}}{1}
      {1}{{{\color{+}1}}}{1}
      {2}{{{\color{+}2}}}{1}
      {3}{{{\color{+}3}}}{1}
      {4}{{{\color{+}4}}}{1}
      {5}{{{\color{+}5}}}{1}
      {6}{{{\color{+}6}}}{1}
      {7}{{{\color{+}7}}}{1}
      {8}{{{\color{+}8}}}{1}
      {9}{{{\color{+}9}}}{1}
      {:}{{{\color{black}{:}}}}{1}
      {,}{{{\color{black}{,}}}}{1}
      {\{}{{{\color{black}{\{}}}}{1}
      {\}}{{{\color{black}{\}}}}}{1}
      {[}{{{\color{black}{[}}}}{1}
      {]}{{{\color{black}{]}}}}{1},
}

\newcommand{\cmark}{\textcolor{teal}{\checkmark}}
\newcommand{\xmark}{\textcolor{red!70!black}{\ensuremath{\times}}}
\newcommand{\pmark}{\textcolor{orange!90!black}{\ensuremath{\circ}}}

\definecolor{+}{RGB}{0,128,0} 
\definecolor{-}{RGB}{139,0,0}

\title{EasyOPD: An Easy-to-use On-Policy Distillation Framework \\ for Large Language Models}

\author{
    Jie Sun$^{1,2,*,\S}$,
    Mao Zheng$^{2,*}$,
    Mingyang Song$^{2,*}$, 
    Qiyong Zhong$^{2,*,\S}$ \\
    \textbf{Gengsheng Li$^{2,*,\S}$,
    Zhepei Hong,
    Chang Wu,
    Pengfei Liu$^1$} \\
    \textbf{Junfeng Fang$^{3,\dagger}$,
    Xiang Wang$^{3,\dagger}$} \\
    $^1$ Shanghai Innovation Institute \quad $^2$ Foundation Model Department, Tencent \\
    $^3$  National University of Singapore \\
    \faGithub\ \href{https://github.com/lds-ustc/EasyOPD}{\textcolor[rgb]{0.0,0.1,0.6}{\texttt{https://github.com/lds-ustc/EasyOPD}}}
}

\begin{document}
\clubpenalty=10000
\widowpenalty=10000
\displaywidowpenalty=10000
\maketitle

\begingroup
\renewcommand{\thefootnote}{}
\NoHyper
\footnotetext{
$^{*}$\,Equal Contribution.\quad
$^{\dagger}$\,Corresponding Authors.\quad
$^{\S}$\,Work done during internship at Tencent.}
\endNoHyper
\endgroup

\input{chapters/0-abs}
\input{chapters/1-intro}
\input{chapters/2-background}
\input{chapters/3-design}
\input{chapters/4-experiment}
\input{chapters/5-conclusion}

\newpage
\bibliography{main}

\newpage
\appendix
\input{chapters/appendix}

\end{document}

%% file: chapters/0-abs.tex
\begin{abstract}
Conventional language-model distillation often relies on fixed teacher-generated data, which may not cover the states encountered by an evolving student policy.
On-policy distillation (OPD) instead collects teacher or evaluator supervision on student-generated rollouts.
However, existing OPD methods differ substantially in supervision form, tokenizer compatibility, teacher access, and supervision granularity, leading to fragmented implementations that are difficult to reproduce and extend.
We present \textsc{EasyOPD}, an on-policy distillation framework built on verl, a distributed reinforcement-learning framework for large language models.
\textsc{EasyOPD} separates user-side configuration, method-specific supervision logic, and verl-based execution.
Its method modules connect to the shared backend through extension boundaries for loss construction, rollout metadata, reward processing, tokenizer alignment, and teacher-side computation.
We instantiate representative methods for three OPD settings---cross-tokenizer OPD, on-policy self-distillation, and step-wise OPD.
Experiments on reasoning, code-generation, scientific-knowledge, and tool-use benchmarks show that these implementations can be executed through the same verl-based backend while retaining their method-specific objectives and task-dependent performance profiles.
We release \textsc{EasyOPD} with runnable YAML configurations, documentation, and an installable demonstration package and video.

\end{abstract}

%% file: chapters/1-intro.tex





\section{Introduction}

Large language model distillation transfers capabilities from a stronger teacher to a smaller, more efficient student, aiming to reduce deployment costs while retaining reasoning, coding, and instruction-following abilities~\citep{alkhulaifi2021knowledge_kd}.
Conventional distillation uses teacher responses or token distributions on fixed contexts~\citep{yang2024kd_survey}.
At inference, however, the student conditions on its own prefixes and may encounter contexts poorly covered by offline supervision~\citep{gu2024minillm_opd,ko2024distillm_opd}.
On-policy distillation (OPD) mitigates this mismatch through a common loop: the current student generates rollouts, a teacher or evaluator supervises them, and the resulting signals guide student updates~\citep{song2026survey,lu2025onpolicydistillation}.
OPD methods vary by supervision interface, tokenizer compatibility, and granularity, with overlapping settings including logit-based distillation, cross-tokenizer alignment~\citep{sun2026simctrecoveringlostsupervision}, on-policy self-distillation, black-box or rubric-based feedback, and step-wise or multi-turn agentic supervision~\citep{zhong2026sod,li2026curriculumturn}.

\begin{figure*}[t]
    \centering
    \includegraphics[width=1.0\textwidth]{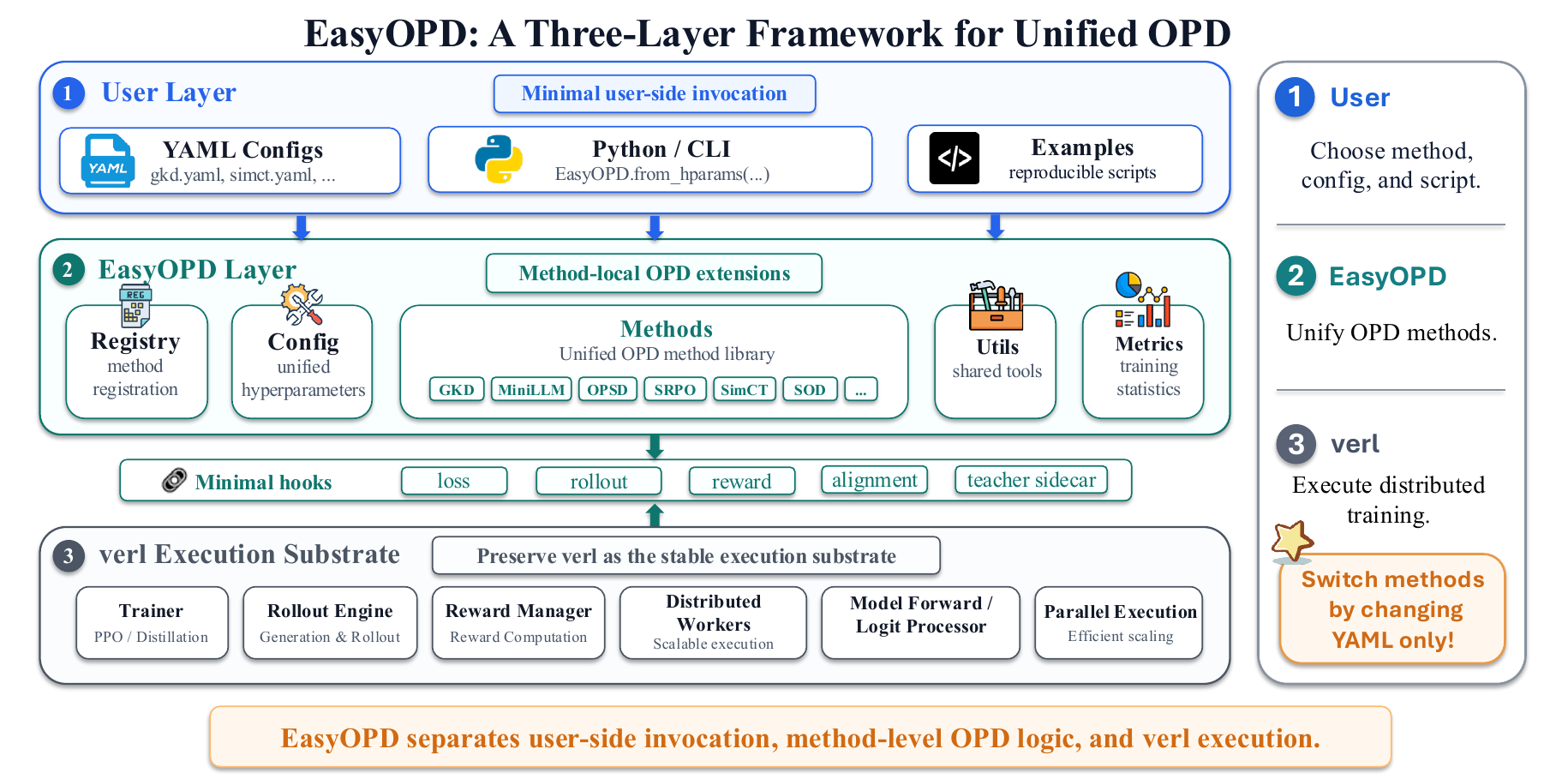}
    \caption{Architecture of \textsc{EasyOPD}. The user layer provides configuration-based invocation, the \textsc{EasyOPD} layer encapsulates method-local OPD logic, and verl provides distributed execution. Hooks for loss, rollout, reward, alignment, and teacher-side computation connect method modules to verl, allowing supported methods to share the same execution backend and be selected via YAML configurations.}
    \label{fig:framework}
\end{figure*}

A shared algorithmic paradigm does not yield a common system interface. Depending on the supervision, OPD may span rollout generation, teacher inference, cross-tokenizer alignment, reward construction, distributed data flow, and optimization. For example, cross-tokenizer OPD requires alignment across teacher and student tokenizations before constructing compatible supervision, whereas black-box or rubric-based OPD may convert textual judgments or scores into rewards; these patterns require different extension points. Existing frameworks address complementary parts of the OPD workflow: TRL~\citep{vonwerra2020trl} provides trainer-centric distillation APIs; verl~\citep{sheng2024hybridflow_verl} and slime~\citep{slime_github} support scalable on-policy rollout and distributed execution; and KDFlow~\citep{zhang2026kdflow} provides a distillation-oriented stack. However, as summarized in Table~\ref{tab:opd-frameworks}, we find no single reviewed framework that documents method-local extension boundaries across the heterogeneous supervision interfaces considered here while reusing a common distributed backend. This gap calls for a lightweight OPD abstraction linking method-specific supervision to a shared execution substrate.

To fill this gap, we present \textsc{EasyOPD}, a unified OPD framework built on verl.
\textsc{EasyOPD} reuses verl for rollout generation, model execution, and optimization while adding a lightweight abstraction layer for method-local supervision.
Method modules connect to the shared training flow through explicit extension points: cross-tokenizer methods localize alignment and supervision construction, whereas black-box or rubric-based methods localize feedback acquisition and training-signal construction.
This design turns OPD integration from framework-wide modification into method-local extension.

To realize this design, \textsc{EasyOPD} adopts the three-layer architecture shown in Figure~\ref{fig:framework}. The user layer provides Python and CLI entry points and YAML configurations for methods, models, datasets, and hyperparameters; the \textsc{EasyOPD} layer manages method registration, method-local modules, and supervision diagnostics, connecting to verl through dispatch hooks for loss, rollout metadata, reward, alignment, and teacher-side computation; and the verl layer handles distributed rollout, model execution, worker orchestration, reward computation, and optimization. Users select methods through configuration, while developers add a method by registering a method-local module with the required hooks, so that the core trainer and workers reach method-specific supervision only through the shared dispatch layer.

We instantiate this design with representative methods across three settings: cross-tokenizer OPD, on-policy self-distillation, and step-wise OPD; its extension boundaries are further designed to accommodate additional supervision forms such as black-box or rubric-based feedback. Within each setting, the compared methods share the model initialization, training data, execution backend, and evaluation procedure where applicable, while retaining their original supervision designs. We release the code under the Apache License 2.0 on GitHub with reproducible YAML configurations, documentation, an installable demonstration package, and a demonstration video.

%% file: chapters/2-background.tex
\section{Background and Motivation}

\paragraph{Previous Solutions.}
Knowledge distillation (KD) typically trains a smaller student model to imitate the predictive behavior of a larger or more capable teacher~\citep{hinton2015distilling_kd,ko2024distillm_opd}. For autoregressive language models, common offline approaches include sequence-level distillation, in which the student is trained on teacher-generated sequences~\citep{kim2016sequence}, and supervised token-level KD, in which the student matches the teacher's next-token distributions along prefixes induced by a fixed offline set of output sequences~\citep{agarwal2024onpolicy_opsd}. These output sequences may be ground-truth or teacher-generated. In their purely offline form, both approaches optimize over an output-prefix distribution that does not track the current student's rollout distribution. During autoregressive inference, however, the student conditions on its own previous predictions. A deviation can therefore shift subsequent prefixes away from those observed during offline distillation, creating a training--inference distribution mismatch and allowing errors to compound across decoding steps~\citep{lin2020imitkd,agarwal2024onpolicy_opsd}.

\paragraph{On-Policy Distillation Mechanism.}
On-policy distillation (OPD) collects teacher supervision on outputs generated by the current student, rather than relying only on fixed offline sequences. During training, the student generates responses or trajectories, and the teacher provides supervision on the resulting prefixes or states. The student is then updated, and new rollouts are generated with the updated policy. In a common white-box setting, the teacher provides next-token distributions on student-generated prefixes, and the student minimizes a token-level divergence from the teacher~\citep{agarwal2024onpolicy_opsd}. The defining property of OPD is therefore that supervision is collected on states induced by the current student, rather than the use of any particular distillation objective~\citep{lu2025onpolicydistillation,song2026survey}.

Training on student-generated states helps reduce the mismatch between offline training data and the prefixes encountered during autoregressive inference, while providing guidance on errors made by the student itself~\citep{agarwal2024onpolicy_opsd,ko2024distillm_opd}. However, OPD also turns distillation into an iterative pipeline that must coordinate student rollout, teacher inference, supervision construction, and model optimization. Since OPD methods differ in teacher access, feedback form, and training objectives~\citep{song2026survey}, a practical framework should support this shared pipeline while allowing method-specific supervision to be implemented consistently.

%% file: chapters/3-design.tex
\section{Design and Implementation}

\input{tables/property_main}
\input{chapters/fig-running-example}

EasyOPD provides a unified OPD training process on top of verl. Instead of treating each OPD method as an isolated training script, it decomposes an OPD run into four reusable stages: method invocation, configuration resolution, supervision construction, and distributed execution. This section describes how these stages form a unified workflow that connects diverse OPD methods to the same verl execution substrate.

\subsection{A Unified OPD Workflow}

Although OPD methods differ in their supervision interfaces, they share a common training flow. Given a teacher model, a student model, and training data, the student first generates on-policy rollouts; the selected method then constructs supervision on these rollouts, such as teacher logits, aligned token distributions, judge rewards, or step-level feedback; finally, the training system consumes this supervision through loss computation, reward aggregation, or trajectory-level weighting. EasyOPD makes this shared process explicit and standardizes invocation and dispatch, without requiring all methods to share a single loss, teacher interface, or evaluation metric. As shown in Figure~\ref{fig:framework}, the user layer specifies what to run, the EasyOPD layer defines how supervision is constructed, and the verl layer executes the resulting training process on top of its distributed runtime~\citep{sheng2024hybridflow_verl}.

\subsection{Declarative Invocation and Configuration}

EasyOPD exposes OPD methods through declarative configurations and launch scripts. Released methods can be selected through method-specific YAML configurations without manually editing rollout workers, reward managers, logit processors, or trainer-side loss assembly, which turns method selection into a configuration-level operation and makes experiments easier to reproduce. Figure~\ref{fig:running_example} shows a concrete running example in which the same \texttt{from\_hparams} entry point and a single launch command drive all three studied settings---cross-tokenizer OPD, on-policy self-distillation, and step-wise OPD---together with their baselines, changing only the method name and its YAML configuration.

Different OPD methods require different options: logit-based methods may specify divergence directions or top-$k$ support; cross-tokenizer methods may specify alignment or overlap-vocabulary behavior; rubric-based methods may specify judge prompts or reward aggregation rules; step-wise methods may specify trajectory segmentation or weighting strategies. EasyOPD keeps these choices in a unified configuration space and resolves them through method registration, so new methods can be added without hard-coding method choices into the trainer.

\subsection{Method-Local Supervision Logic}

The core of EasyOPD is to localize OPD-specific supervision logic: a method defines how supervision is obtained, transformed, consumed, and diagnosed. Standard OPD may only require a token-level distillation objective, cross-tokenizer OPD may require tokenizer or span alignment and a common supervision space, rubric-based OPD may convert judge feedback into rewards, and step-wise OPD may attach feedback to intermediate reasoning states or multi-turn trajectories. EasyOPD does not force these methods into a single loss-only abstraction; it gives each a local space to implement the supervision interface it needs.

Method-local logic also includes diagnostics. Implemented method modules expose supervision-specific diagnostics in addition to task performance, so users can check whether the intended supervision signal is active, not only the final loss. For example, the released cross-tokenizer implementation reports the overlap-vocabulary size, the number of valid and span-level aligned segments, the count of skipped samples, and the valid-span ratio, while the step-wise implementation reports the mean step-level weight and the number of supervised steps. These metrics help users debug failures that final task performance alone cannot explain.

\subsection{Execution through Thin verl Hooks}

EasyOPD reuses verl as the execution substrate instead of reimplementing distributed training. verl provides the system-heavy components required by OPD, including rollout generation, reward computation, model forwarding, logit processing, distributed workers, trainer-side optimization, and parallel execution, and EasyOPD connects method-local supervision logic to this substrate through a small set of thin hooks.

A hook is a dispatch boundary rather than a method implementation: it exposes a stable interaction point in the training pipeline and routes the corresponding data or control flow to the selected method module. The hook set covers the main entry points of OPD supervision: loss hooks for token- or distribution-level supervision, rollout hooks for trajectory metadata, reward hooks for judge or rubric feedback, alignment hooks for token- or span-level mapping, and teacher-sidecar hooks for teacher-side signals such as logits, hidden states, or judge outputs, as used respectively by logit-based, feature-based~\citep{zhang2024dskd}, and rubric-based methods.

This design balances expressiveness and maintainability. Writing every method directly into verl would create method-specific branches, while exposing only a single fixed interface would make alignment-, reward-, or trajectory-based OPD methods difficult to express. Thin hooks keep method logic local while verl remains the shared backend. EasyOPD thus inherits rollout generation, worker orchestration, optimization, and distributed execution from verl; its contribution is the dispatch layer connecting these components to method-local supervision logic.

%% file: tables/property_main.tex
\begin{table*}[t]
\centering
\small
\caption{
Comparison of OPD supervision interfaces and framework properties.
\cmark\ denotes native or first-class support, \pmark\ denotes partial
or experimental support, and \xmark\ denotes no documented support.
\emph{On-policy self-distillation} refers to settings in which the
teacher is derived from the student model and supervises
student-generated rollouts. The comparison reflects publicly documented
functionality as of July 2026, using the versions or commits
listed in Appendix~\ref{app:framework-comparison}, where per-cell evidence
is also provided.
}
\setlength{\tabcolsep}{6pt}
\begin{tabular}{l cccc ccc}
\toprule
 & \multicolumn{4}{c}{\textbf{OPD supervision-interface coverage}}
 & \multicolumn{3}{c}{\textbf{Framework properties}} \\
\cmidrule(lr){2-5}\cmidrule(lr){6-8}

\textbf{Framework}
 & \shortstack{~\\OPD}
 & \shortstack{Cross-tok.\\OPD}
 & \shortstack{On-policy\\self-distill.}
 & \shortstack{Step-wise\\OPD}
 & \shortstack{Released\\YAML/CLI}
 & \shortstack{Documented method\\extension API}
 & \shortstack{Compatible w/\\multi-node} \\
\midrule

TRL
 & \cmark
 & \pmark
 & \pmark
 & \xmark
 & \cmark
 & \xmark
 & \pmark \\

verl
 & \cmark
 & \xmark
 & \xmark
 & \pmark
 & \cmark
 & \xmark
 & \cmark \\

slime
 & \cmark
 & \xmark
 & \pmark
 & \pmark
 & \cmark
 & \xmark
 & \cmark \\

KDFlow
 & \cmark
 & \cmark
 & \cmark
 & \xmark
 & \cmark
 & \pmark
 & \pmark \\

\textbf{Ours}
 & \cmark
 & \cmark
 & \cmark
 & \cmark
 & \cmark
 & \cmark
 & \cmark\,{\footnotesize(via verl)} \\

\bottomrule
\end{tabular}

\label{tab:opd-frameworks}
\end{table*}

%% file: chapters/fig-running-example.tex
\definecolor{cmtgray}{HTML}{6E7781}
\definecolor{kwblue}{HTML}{0B5FA5}
\definecolor{strgreen}{HTML}{0B6E4F}
\lstdefinestyle{easyopd}{
    basicstyle=\ttfamily\scriptsize,
    commentstyle=\color{cmtgray},
    keywordstyle=\color{kwblue}\bfseries,
    stringstyle=\color{strgreen},
    numbers=none,
    frame=none,
    breaklines=true,
    columns=fullflexible,
    keepspaces=true,
    showstringspaces=false,
    aboveskip=1pt, belowskip=1pt,
}

\begin{figure*}[t]
    \centering
    \begin{tcolorbox}[colback=black!2, colframe=black!45, boxrule=0.6pt,
                      arc=3pt, left=5pt, right=5pt, top=2pt, bottom=2pt,
                      title=\footnotesize\textbf{One entry point for every OPD setting in \textsc{EasyOPD}}]
    \begin{lstlisting}[style=easyopd,language=Python,
        morekeywords={from,import,print}]
from easyopd import EasyOPD
print(EasyOPD.list_methods())   # [..., 'simct','uld','alm','dskd','sdpo','sod', ...]
# The SAME call selects any method; only the name + YAML change.
m = EasyOPD.from_hparams("simct", config_path="easyopd/config/simct.yaml")  # cross-tokenizer OPD (uld|alm|dskd)
m = EasyOPD.from_hparams("sdpo",  config_path="easyopd/config/sdpo.yaml")   # on-policy self-distillation (vs grpo)
m = EasyOPD.from_hparams("sod",   config_path="easyopd/config/sod.yaml")    # step-wise OPD (vs response-level opd, grpo)
    \end{lstlisting}
    \hrule height 0.3pt
    \begin{lstlisting}[style=easyopd,language=bash]
# Each method (and its baselines) is one launch script; the config selects the method and the hooks it activates.
$ bash examples/simct/run_simct.sh   # cross-tokenizer: uld|alm|dskd
$ bash examples/sdpo/run_sdpo.sh     # self-distillation: grpo
$ bash examples/sod/run_sod.sh       # step-wise: response-level opd, grpo
    \end{lstlisting}
    \hrule height 0.3pt
    \begin{lstlisting}[style=easyopd,language=]
# The registry activates only the hooks a method needs and logs supervision-specific diagnostics:
#   simct -> alignment + teacher sidecar + loss  : simct/xtok_kd_loss, simple/teacher_loss_tokens_mean
#   sdpo  -> reprompt self-teacher + loss        : actor/sdpo/loss, self_distillation/reprompt_sample_fraction
#   sod   -> step-wise KL reweighting            : actor/token_kl, stepwise_opd_coef
    \end{lstlisting}
    \end{tcolorbox}
    \caption{A running example of \textsc{EasyOPD}. The three studied settings---cross-tokenizer OPD, on-policy self-distillation, and step-wise OPD---and their baselines are all reached through the \emph{same} \texttt{from\_hparams} entry point or a single launch script, changing only the method name and its YAML config. The registry then activates only the hooks each method needs, verl performs rollout and optimization, and \textsc{EasyOPD} reports supervision-specific diagnostics. All API calls, method names, launch commands, and diagnostic metric names are taken from the released repository.}
    \label{fig:running_example}
\end{figure*}

%% file: chapters/4-experiment.tex
\section{Experiments}
\label{sec:experiments}

\subsection{Experimental Setup}
\label{sec:exp_setup}

We evaluate \textsc{EasyOPD} in three settings that exercise different method-specific interfaces. Cross-tokenizer OPD requires supervision across different teacher and student tokenizations. On-policy self-distillation instantiates teacher and student policies from the same model under different conditioning contexts. Step-wise OPD assigns supervision to intermediate reasoning or tool-interaction steps. These settings cover variation in supervision space, teacher conditioning, and supervision granularity, and are not intended to form a single cross-regime leaderboard across methods.

\paragraph{Cross-tokenizer OPD.}
Standard OPD directly compares teacher and student next-token distributions, which are not naturally aligned under different tokenizers; cross-tokenizer methods address this by constructing a shared supervision space. We compare SimCT \citep{sun2026simctrecoveringlostsupervision}, ULD \citep{boizard2025uld}, ALM \citep{minixhofer2025alm}, and DSKD \citep{zhang2024dskd}, distilling Qwen2.5-7B-Instruct \citep{qwen25} into Phi-4-mini-Instruct \citep{phi4mini}. All methods start from the same student checkpoint, warm-started on a shared 10K teacher-generated math and code corpus, and use the same training prompts and optimization budget. We evaluate on MATH500 \citep{lightman2023math500}, GSM8K~\citep{cobbe2021training_gms8k}, MBPP~\citep{austin2021program_mbpp}, and LiveCodeBench v6~\citep{jain2024livecodebench}, reporting exact-match accuracy for reasoning and pass@1 for code.

\paragraph{On-policy self-distillation.}
On-policy self-distillation (OPSD) uses the same base model under different conditioning contexts: the student generates a response from the original prompt, while the self-teacher re-evaluates it with additional feedback. We instantiate this setting with SDPO~\citep{hubotter2026sdpo}, using Qwen3-8B~\citep{qwen2025qwen3} as both the student and self-teacher, and compare it with the base model and GRPO~\citep{shao2024deepseekmath}. We evaluate Chemistry from SciKnowEval~\citep{feng2024sciknoweval}, Biology from SciKnowEval~\citep{feng2024sciknoweval}, Tool Use from ToolAlpaca~\citep{tang2023toolalpaca}, and mathematical reasoning on GSM8K~\citep{cobbe2021training_gms8k}. We report the highest Average@16 reached within one-hour and five-hour wall-clock budgets, where Average@16 is the mean correctness over 16 independently sampled responses per question.

\paragraph{Step-wise OPD.}
Response-level OPD~\citep{agarwal2024onpolicy_opsd} applies teacher supervision uniformly across a trajectory, although supervision on later steps may become less reliable after erroneous tool interactions, a common failure mode as LLMs are deployed as long-horizon, tool-use, and research agents~\citep{xiao2025limi,li2026agencybench,feng2026longclibench,wu2025innovatorbench,fu2026argo,zeng2026davincidev,fu2026davincienv}. SOD~\citep{zhong2026sod} instead reweights the distillation loss by step-level student--teacher divergence. We use Qwen3-1.7B~\citep{qwen2025qwen3} as the student and a GRPO-optimized Qwen3-4B~\citep{zhong2026sod} as the teacher, comparing the base model, GRPO~\citep{shao2024deepseekmath}, response-level OPD, and SOD under a shared SFT initialization, training data, and optimization budget \citep{zhong2026sod}. We evaluate on AIME 2024, AIME 2025~\citep{maa_aime}, and LiveCodeBench v6~\citep{jain2024livecodebench}; each score is Average@32 over 32 sampled responses per problem, and Avg.\ is their unweighted mean.

\paragraph{Shared implementation.}
All methods are implemented in \textsc{EasyOPD} and run on the same verl-based infrastructure~\citep{sheng2024hybridflow_verl}. Within each setting we fix the model, training data, backend, and evaluation procedure where applicable, retaining each method's original supervision design. Optimization, sampling, hardware, and checkpoint details are in Appendix~\ref{app:exp_setup}.

\subsection{Results}
\label{sec:exp_results}

\paragraph{Cross-tokenizer OPD.}
Table~\ref{tab:cross_tokenizer_results} shows that all four evaluated cross-tokenizer methods improve the unweighted average over the base student. SimCT obtains the highest average (53.5, +3.3 over the base) and the best MATH500, ALM leads on MBPP and LiveCodeBench v6, and DSKD is best on GSM8K. This confirms that the evaluated cross-tokenizer objectives run through \textsc{EasyOPD}'s alignment and loss interfaces under a matched setup, while retaining distinct task-dependent profiles.

\input{tables/cross-tokenizer}
\input{tables/opsd}
\input{tables/agentic-opd}

\paragraph{On-policy self-distillation.}
Table~\ref{tab:self_opd_results} compares SDPO and GRPO under the two wall-clock budgets. SDPO's gains are largest on Chemistry and Biology (+19.2 points at one hour, +6.7 and +6.0 at five hours), while GRPO is stronger on GSM8K and on the five-hour Tool Use result. SDPO's relative performance is thus task-dependent in the evaluated setting.

\paragraph{Step-wise OPD.}
Table~\ref{tab:stepwise_opd_results} shows that response-level OPD exceeds GRPO on all three benchmarks, and SOD obtains the highest observed score on each, for an unweighted average of 39.95 (+5.1 over response-level OPD). These results are consistent with the intended role of step-wise weighting, but do not by themselves establish that every step-wise OPD method will outperform response-level OPD.

\paragraph{Overall findings.}
Across the three case studies, \textsc{EasyOPD} runs the evaluated methods on the same verl backend while preserving their distinct supervision requirements; results should be read within each setting rather than as a cross-regime ranking.

%% file: tables/cross-tokenizer.tex
\begin{table}[t]
    \centering
    \small
    \setlength{\tabcolsep}{4pt}
    \renewcommand{\arraystretch}{1.15}
    \caption{Cross-tokenizer OPD results. Reasoning scores are exact-match accuracy, code scores are pass@1, and Avg.\ is their unweighted mean. Best per column in bold.}
    \label{tab:cross_tokenizer_results}
    \begin{tabular}{lccccc}
        \toprule
        Method
        & MATH500
        & GSM8K
        & MBPP
        & LCB-v6
        & Avg. \\

        \midrule

        Base
        & 54.2
        & 86.0
        & 53.2
        & 7.4
        & 50.2 \\

        \midrule

        ULD
        & 57.0
        & 85.1
        & 55.4
        & 9.1
        & 51.7 \\

        ALM
        & 58.2
        & 84.9
        & \textbf{56.4}
        & \textbf{12.0}
        & 52.9 \\

        DSKD
        & 58.8
        & 86.4
        & 56.2
        & 9.7
        & 52.8 \\

        SimCT
        & \textbf{59.6}
        & \textbf{86.7}
        & \textbf{56.4}
        & 11.4
        & \textbf{53.5} \\

        \bottomrule
    \end{tabular}
\end{table}

%% file: tables/opsd.tex
\begin{table}[t]
    \centering
    \small
    \setlength{\tabcolsep}{3.5pt}
    \renewcommand{\arraystretch}{1.15}
    \caption{
    SDPO and GRPO in the OPSD setting under one-hour and five-hour wall-clock
budgets. Average@16 is the mean correctness over 16 sampled responses per
question; Base is the untrained checkpoint, reported once per task. We evaluate
every five steps and report the best Average@16 observed within each budget;
the best per task and budget is in bold.
    }
    \label{tab:self_opd_results}
    \begin{tabular}{lcccccccc}
        \toprule
        & \multicolumn{8}{c}{$\mathrm{avg@16}$} \\
        \cmidrule(lr){2-9}
        Method
        & \multicolumn{2}{c}{Chemistry}
        & \multicolumn{2}{c}{Biology}
        & \multicolumn{2}{c}{Tool Use}
        & \multicolumn{2}{c}{GSM8K} \\
        \cmidrule(lr){2-3}
        \cmidrule(lr){4-5}
        \cmidrule(lr){6-7}
        \cmidrule(lr){8-9}
        & 1h & 5h
        & 1h & 5h
        & 1h & 5h
        & 1h & 5h \\
        \midrule
        Base
        & \multicolumn{2}{c}{41.1}
        & \multicolumn{2}{c}{30.5}
        & \multicolumn{2}{c}{57.7}
        & \multicolumn{2}{c}{86.1} \\

        \midrule
        
        GRPO
        & 51.2 & 72.0
        & 32.1 & 53.9
        & 62.3 & \textbf{65.8}
        & \textbf{91.8} & \textbf{93.3} \\

        SDPO
        & \textbf{70.4} & \textbf{78.7}
        & \textbf{51.3} & \textbf{59.9}
        & \textbf{62.8} & 62.8
        & 87.7 & 87.7 \\
        \bottomrule
    \end{tabular}
\end{table}

%% file: tables/agentic-opd.tex
\begin{table}[t]
    \centering
    \small
    \setlength{\tabcolsep}{7pt}
    \renewcommand{\arraystretch}{1.15}
    \caption{
    Step-wise OPD with Qwen3-1.7B as the student. OPD and SOD use a GRPO-optimized Qwen3-4B teacher. Values are percentages; Avg.\ is the unweighted mean across AIME 2024, AIME 2025, and LiveCodeBench v6. Best per column in bold.
    }
    \label{tab:stepwise_opd_results}
    \begin{tabular}{lcccc}
        \toprule
        Method
        & AIME24
        & AIME25
        & LCB-v6
        & Avg. \\
        \midrule
        Base
        & 17.19
        & 12.81
        & 12.41
        & 14.14 \\

        \midrule
        
        GRPO
        & 34.06
        & 26.25
        & 21.93
        & 27.41 \\

        OPD
        & 38.13
        & 36.04
        & 30.27
        & 34.81 \\

        SOD
        & \textbf{43.65}
        & \textbf{38.33}
        & \textbf{37.88}
        & \textbf{39.95} \\
        \bottomrule
    \end{tabular}
\end{table}

%% file: chapters/5-conclusion.tex
\section{Conclusion and Future Work}

We present EasyOPD, a method-oriented framework that integrates on-policy distillation methods on a shared verl backend, separating user-side configuration, method-local supervision logic, and distributed execution through thin dispatch hooks. Across cross-tokenizer OPD, on-policy self-distillation, and step-wise OPD, methods with different supervision interfaces share this backend while keeping their own objectives and diagnostics; future work will broaden method coverage and larger-model evaluation.

%% file: chapters/appendix.tex
\section{Framework Comparison Details}
\label{app:framework-comparison}

\input{tables/property_apd}
Table~\ref{tab:opd-frameworks-detailed} provides the
framework-specific evidence underlying the ratings in
Table~\ref{tab:opd-frameworks}.

\section{Experimental Setup Details}
\label{app:exp_setup}

We summarize the optimization, sampling, and hardware settings for the three
case studies. Unless noted otherwise, all settings are taken from the released
configuration files and launch scripts.

\paragraph{Software environment.}
All experiments run on a node with $8\times$ NVIDIA H20 96GB GPUs
(driver 535.161.08, CUDA 12.4) under Ubuntu 22.04, using the released
\texttt{OpenAgentRL} conda environment (Python 3.11, PyTorch 2.6.0+cu124,
verl 0.5.0, Ray 2.47.1, Transformers 4.51.1, vLLM 0.8.5.post1,
SGLang 0.4.6.post1, FlashAttention 2.7.4.post1, and xFormers 0.0.29.post2).
The full pinned dependency list and a one-click installer are released with the
code.

\paragraph{Cross-tokenizer OPD.}
We distill Qwen2.5-7B-Instruct (vocabulary size 151{,}936) into a
Phi-4-mini-Instruct student (3.8B, vocabulary size 200{,}064) that is first
SFT-warm-started on a shared 10K mixed math-and-code corpus derived from the
single-turn prompt--answer pairs of the \texttt{mixed\_math\_code\_10k} dataset.
Training uses GRPO with a cross-tokenizer KL reward, a learning rate of
$5\times10^{-7}$ with a $0.05$ warmup ratio, a train batch size of $64$, one
epoch ($154$ steps over the $\sim$9.9K training prompts), maximum prompt and
response lengths of $4096$ tokens, and single-sample rollouts at temperature
$0.6$. All compared methods share this training-prompt set and optimization
budget. We save a checkpoint every $20$ steps and, after merging, evaluate each
checkpoint with an SGLang-based inference server, reporting exact-match accuracy
for reasoning (verified with \texttt{math\_verify}) and pass@1 for code on
MATH500, GSM8K, MBPP, and LiveCodeBench v6.

\paragraph{On-policy self-distillation.}
SDPO uses Qwen3-8B as both the student and the self-teacher, with an EMA
self-teacher (update rate $0.05$), top-$k$ logit distillation with $k=100$, and
a symmetric KL interpolation ($\alpha=0.5$). Following the SDPO
setup~\citep{hubotter2026sdpo}, we evaluate a checkpoint every five training
steps and report the best Average@16 observed within the one-hour and five-hour
wall-clock budgets, where Average@16 is the mean correctness over 16
independently sampled responses per question.

\paragraph{Step-wise OPD.}
We use Qwen3-1.7B as the student and a GRPO-optimized Qwen3-4B as the teacher,
with a shared SFT initialization, training data, and optimization budget across
compared methods. Rollouts are multi-turn tool-interaction trajectories, and
SOD applies step-wise adaptive weighting (with numerical-stability offset
$\epsilon=10^{-6}$ and bound offset $\delta=0.2$) on top of the response-level
OPD objective. Each benchmark score is Average@32 over 32 independently sampled
responses per problem.

\section{Released Methods and Hooks}
\label{app:method-hooks}

Table~\ref{tab:method-hooks} lists the released methods used in this work and
the dispatch hooks each one registers. Every method generates on-policy
rollouts through verl's shared rollout engine; the hooks in the table are the
\emph{additional} method-local extension points a method registers on top of
this shared pipeline. Because a hook only routes data or control to a method
module, a method activates only the hooks it needs: a logit-based objective
such as GKD registers a single loss hook, whereas methods that require
teacher-side computation or per-step trajectory metadata additionally register
a teacher-sidecar or rollout-metadata hook. All other pipeline stages,
including rollout generation, worker orchestration, and optimization, are
reused from verl unchanged.

\begin{table}[t]
    \centering
    \small
    \setlength{\tabcolsep}{5pt}
    \renewcommand{\arraystretch}{1.15}
    \caption{Method-local dispatch hooks registered by each released method,
    beyond the shared rollout--optimization pipeline that all methods reuse from
    verl. \cmark\ marks an activated hook; the rollout-metadata hook attaches
    per-step or trajectory annotations and is distinct from rollout generation
    itself. Cross-tokenizer alignment in SimCT reuses the shared \texttt{simple}
    teacher sidecar and is applied inside its loss path.}
    \label{tab:method-hooks}
    \begin{tabular}{lcccc}
        \toprule
        Method & Loss & \shortstack{Rollout\\meta.} & \shortstack{Teacher\\sidecar} & Setting \\
        \midrule
        GKD   & \cmark & \xmark & \xmark & Logit \\
        SimCT & \cmark & \xmark & \cmark & Cross-tok. \\
        SDPO  & \cmark & \xmark & \cmark & Self-distill. \\
        SOD   & \cmark & \cmark & \xmark & Step-wise \\
        \bottomrule
    \end{tabular}
\end{table}

%% file: tables/property_apd.tex
\begin{table*}[t]
\centering
\small
\setlength{\tabcolsep}{6pt}
\begin{tabular}{l cccc ccc}
\toprule
 & \multicolumn{4}{c}{\textbf{OPD supervision-interface coverage}}
 & \multicolumn{3}{c}{\textbf{Framework properties}} \\
\cmidrule(lr){2-5}\cmidrule(lr){6-8}

\textbf{Framework}
 & \shortstack{~\\OPD}
 & \shortstack{Cross-tok.\\OPD}
 & \shortstack{On-policy\\self-distill.}
 & \shortstack{Step-wise\\OPD}
 & \shortstack{Released\\YAML/CLI}
 & \shortstack{Documented method\\extension API}
 & \shortstack{Compatible w/\\multi-node} \\
\midrule

TRL
 & \cmark
 & \pmark$^{a}$
 & \pmark$^{b}$
 & \xmark
 & \cmark
 & \xmark$^{c}$
 & \pmark$^{d}$ \\

verl
 & \cmark
 & \xmark
 & \xmark
 & \pmark$^{e}$
 & \cmark
 & \xmark$^{f}$
 & \cmark \\

slime
 & \cmark$^{g}$
 & \xmark
 & \pmark$^{h}$
 & \pmark$^{e,g}$
 & \cmark
 & \xmark$^{f}$
 & \cmark \\

KDFlow
 & \cmark
 & \cmark
 & \cmark$^{i}$
 & \xmark
 & \cmark
 & \pmark$^{j}$
 & \pmark$^{k}$ \\

\textbf{Ours}
 & \cmark
 & \cmark
 & \cmark
 & \cmark
 & \cmark
 & \cmark
 & \cmark \\
\bottomrule
\end{tabular}

\caption{
Framework-specific evidence supporting the ratings in
Table~\ref{tab:opd-frameworks}. Ratings reflect publicly documented
functionality of the following versions: EasyOPD builds on verl~0.5.0,
and we compare against TRL~0.27.0, the slime repository (accessed
2025-06-19), and the KDFlow implementation~\citep{zhang2026kdflow}.
Where a capability could not be verified from public documentation or
code, it is marked as not documented rather than inferred as absent.
}
\label{tab:opd-frameworks-detailed}

\vspace{2pt}
\begin{minipage}{0.99\textwidth}
\footnotesize
\raggedright

$^{a}$ Cross-tokenizer support is provided through experimental
distillation components rather than the standard GKD path.

$^{b}$ TRL provides an experimental SDPO trainer with live, base, and
EMA teacher options, but the support is tied to a specialized trainer.

$^{c}$ OPD variants are implemented as separate trainer classes rather
than composable method-level extension points.

$^{d}$ Multi-node trainer execution is available, but a disaggregated
asynchronous rollout--teacher--trainer pipeline is not provided.

$^{e}$ Agentic or step-wise supervision is expressible through general
reward/RLVR and multi-turn infrastructure, but is not exposed as a
dedicated OPD interface.

$^{f}$ New variants require method-specific logic across internal
rollout, reward/advantage, logit-processing, or loss components.

$^{g}$ slime provides OPD through its RL reward/advantage pipeline and
offers strong agentic and multi-turn rollout infrastructure.

$^{h}$ A released slime example uses the original model as the teacher
for self-distillation, but does not expose a general live or EMA
self-teacher interface.

$^{i}$ KDFlow supports student-to-teacher weight synchronization and
EMA teacher updates for on-policy self-distillation.

$^{j}$ KDFlow provides decoupled algorithm APIs, but they remain tied
to its own distillation stack.

$^{k}$ KDFlow supports distributed execution, while its documented
on-policy training loop is primarily synchronous.

\end{minipage}
\end{table*}